# A Comparative Analysis of Machine Learning Methods for Lane Change Intention Recognition Using Vehicle Trajectory Data

**Renteng Yuan (First author)**
Jiangsu Key Laboratory of Urban ITS
School of Transportation
Southeast University, Nanjing, Jiangsu, P. R. China, and 210000
Email: rtengyuan123@126.com

**Abstract:**

Accurately detecting and predicting lane change (LC)processes can help autonomous vehicles better understand their surrounding environment, recognize potential safety hazards, and improve traffic safety. This paper focuses on LC processes and compares different machine learning methods' performance to recognize LC intention from high-dimensionality time series data. To validate the performance of the proposed models, a total number of 1023 vehicle trajectories is extracted from the CitySim dataset. For LC intention recognition issues, the results indicate that with ninety-eight percent of classification accuracy, ensemble methods reduce the impact of Type II and Type III classification errors. Without sacrificing recognition accuracy, the LightGBM demonstrates a sixfold improvement in model training efficiency than the XGBoost algorithm.

**Introduction**

Lane-changing behavior induces spatiotemporal interactions between vehicles, which have a significant impact on both traffic efficiency and safety. Statistical data shows that lane-changing behaviors are responsible for 18% of all roadway crashes and contribute to 10% of delays in China[1]. In 2015, the National Highway Traffic Safety Administration (NHTSA) reported approximately 451,000 traffic accidents in the US related to lane-changing behavior[2]. Timely identification and prediction of lane-changing behaviors are crucial for reducing accidents, enhancing traffic safety, and optimizing road operations.

Lane-changing behavior is a complex process influenced by various factors, including human, vehicle, road, and environmental factors[3]. Lane change intention is defined as the planned or intended action of a driver to change lanes while driving. In previous studies, various indicators are used to characterize lane-changing intentions, such as vehicle dynamics parameters (e.g., steering wheel angle, rate of steering angle change, brake pedal position, and turn signal status)[4-6], driver's physiological indicators (e.g., eye movements and head rotation angle)[7-9], and vehicle operating state indicators (e.g., speed, acceleration, and headway distance)[10-12]. However, the practical application of vehicle dynamics parameters is limited due to variations in driving habits, affecting the recognition performance of related models. For example, turn signal utilization rates for lane-changing vehicles have been reported to be 44% and 40% in the United States and China, respectively[13, 14]. The acquisition of driver's physiological indicators faces challenges related to experimental conditions, such as small sample sizes and high data homogeneity, which can impact the transferability and reliability of trained models. Monitoring driver's physiological characteristics faces challenges related to data quality, cost, and potential discomfort to the driver. With the development of vehicle-to-vehicle communication and vehicle-to-infrastructure technologies, access to personalized, high-precision vehicle trajectory data has increased. Vehicle operating statue indicators can be extracted directly from the vehicle trajectory and are increasingly used for lane change intention recognition due to their easy accessibility and large sample size[15-17]. Compared to conducting real-world or driving simulator experiments, vehicle trajectory data is more accessible and overcomes limitations related to small sample sizes and data homogeneity. This study focuses on using vehicle trajectory data to identify lane-changing behaviors by considering interactive influences among adjacent vehicles.

From the methodology perspective, lane-changing intention recognition algorithms can be categorized into two main groups: statistical theory-based methods and machine learning approaches. Common statistical methods include multinomial logit regression models[18] and Bayesian networks[5, 19], which offer high interpretability but may produce biased predictions when the collected data deviates from assumed statistical distributions and hypotheses. In recent years, machine learning algorithms have made rapid progress and gained significant attention across various domains. Machine learning methods, such as support vector machines[20], long short-term memory neural networks (LSTM)[10, 17, 21], and ensemble learning methods[22] are employed in lane-changing intention recognition to capture non-linear relationships among parameters. However, previous studies have primarily focused on evaluating the accuracy of the model and have largely ignored the impact of increasing model complexity on the training time. This study conducts a comparative analysis of various machine learning models for vehicle lane change intention recognition, considering their accuracy and training complexity.

The remainder of this paper is organized as follows: Section 2 presents the details of the proposed methodology. Data process is presented in Section 3. Experimental setup and results are provided in Section 4. Finally, Section 5 summarizes the conclusions and limitations.

**Vehicle Trajectory Data**

The NGSIM, CitySim, and HighD trajectory datasets are extensively utilized vehicle trajectory datasets in academic research. Among these datasets, only CitySim provides the coordinates of the vehicle bounding box. Compared to other publicly available trajectory datasets, CitySim employs seven bounding box points to characterize vehicles and is more suitable for fine-grained studies of driving behavior. Therefore, the CitySim dataset was chosen for this study to validate the performance of the model. Vehicle bounding box feature description is shown in Figure 1.

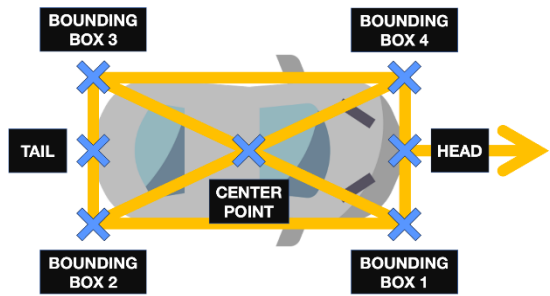

Fig.1. Vehicle bounding box feature description in CitySim[23]

CitySim is a publicly available drone-based vehicle trajectory dataset that contains detailed driving data, vehicle data, and supporting information for the study of driving trajectory and driving intention[23]. A sub-dataset Freeway-B with six lanes in two directions was used in this research. The freeway-B dataset was collected using two UAVs simultaneously over a 2230-ft basic freeway segment. A total of 5623 vehicle trajectories were extracted from 60 minutes of drone videos. Figure 2 displays a snapshot of the freeway-B segment.

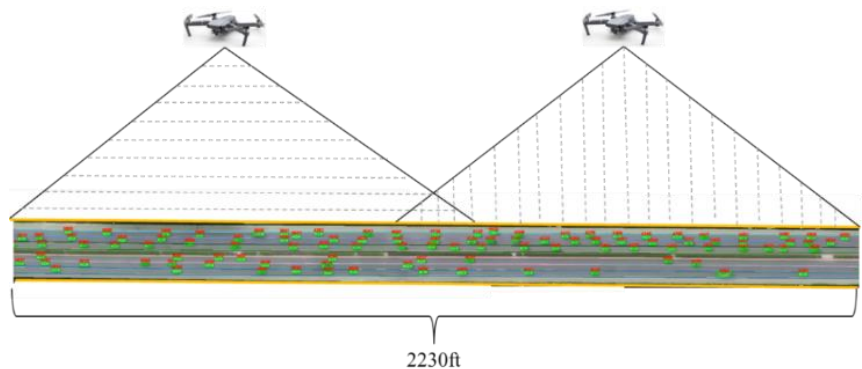

Fig. 2. A snapshot of the freeway-B segment

This study focuses on lane change processes. A total of 1023 vehicle trajectories were extracted from the freeway-B dataset. Among these, 545 trajectories were for lane-change (LC) vehicles, with 240 trajectories for left lane changes (LLC) and 305 trajectories for right lane changes

(RLC). The remaining 478 trajectories were for lane-keeping (LK) vehicles. Lane-keeping vehicle trajectories are randomly extracted.

*Data Processing*

The Freeway-B dataset is derived from two merged drone videos. To minimize the effects of frame misalignment or skipping, vehicle trajectories with variations greater than one in adjacent frames are eliminated. To reduce the negative effect of errors, a moving average (MA) method is used to smooth the trajectory, and the moving average filter is set to 0.5s. A comparison of the original trajectory and processed trajectory is shown in Figure 3.

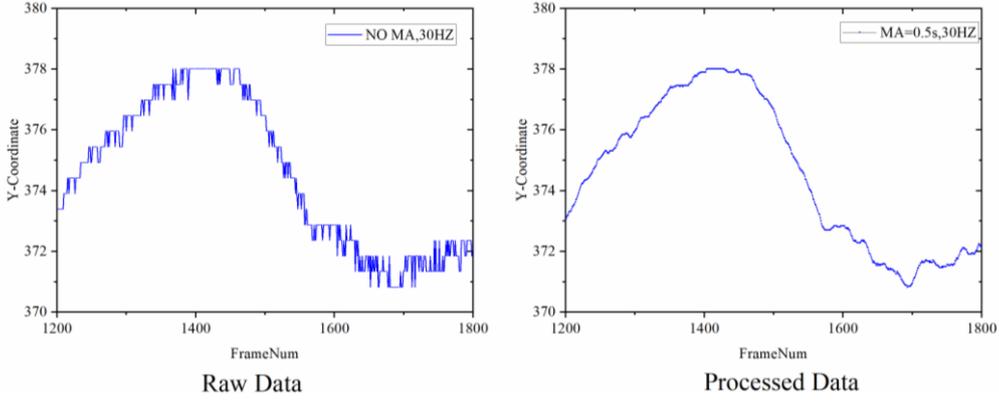

Raw Data            Processed Data

Fig. 3. Comparison of original trajectory and processed trajectory

*Indicator Calculation.*

To accurately describe the vehicle driving status, six indicators are extracted from the two-dimensional (i.e. longitudinal and lateral) vehicle position coordinates, including the longitudinal velocity ($v_x$), lateral velocity ($v_y$), longitudinal acceleration ($a_x$), lateral acceleration ($a_y$), vehicle heading ($\theta$), and yawRate($\triangle \theta$). Furthermore, a non-linear low-pass filter is employed to reduce the negative effect of measurement errors [24]. First, the vehicle speed at the *t-th* frame is calculated and can be formulated as.

$$v_n(t) = \frac{s(t+n) - s(t-n)}{2 \cdot nT} \quad (1)$$

Where *t* represents the current frame, T is a constant, representing 1/30s in this research, *n* represents the time-step; $s(t-n)$ represents the vehicle's position in the frame $t-n$, where *n* takes different values, a vector $\{v_1(t), v_2(t), …, v_N(t)\}$ (In this paper, *n* is set to 8) will be obtained. Thus, the vehicle velocity $v(t)$ at the *t-th* frame is calculated by taking the median of all N time steps. The lateral velocity ($v_y$) and longitudinal velocity ($v_x$) can be determined based on the change in the lateral and longitudinal positions of the vehicle, respectively. With the calculated velocity, acceleration can be obtained as.

$$a(t) = \frac{v(t+1) - v(t-1)}{2 \cdot T} \quad (2)$$

The lateral acceleration ($a_y$) and longitudinal acceleration ($a_x$) also can be determined based on the change of $v_y$ and $v_x$, respectively. In addition, the vehicle heading can be calculated as,

$$\theta_n(t) = \arctan\left(\frac{y_H(t+n) - y_R(t-n)}{x_H(t+n) - x_R(t-n)}\right) \quad (3)$$

Where $\theta_n(t)$ represents the vehicle heading at the frame $t$, $y_H(t+n)$ is the vehicle head point longitudinal position in the frame $t+n$, $x_R(t+n)$ denotes vehicle tail point horizontal position in frame $t+n$. yawRate is used to represent the rate of change of the vehicle's steering wheel angle[25]. It is calculated as,

$$\Delta\theta(t) = \frac{\theta(t+1) - \theta(t-1)}{2T} \quad (4)$$

*Input indicator*

To fully consider the impact of various factors, the input of the combined model consists of three parts: subject vehicle (S-vehicle) information, surrounding vehicle information, and relative position information. Surrounding vehicles include the closest preceding and following vehicles in the adjacent and the current lanes. The ego vehicle is the human-driven vehicle. This research aims to detect and predict the human-driven vehicle LC process. The six indicators ($v_x$, $v_y$, $a_x$, $a_y$, $\theta$, $\Delta\theta$) were calculated for each vehicle. Limited by the video coverage, some trajectory fragments of surrounding vehicles were not recorded. A categorical variable (0 means it has recorded trajectory information; 1 means the trajectory information is missing) is added to each surrounding vehicle indicating this phenomenon. For instance, when the ego vehicle first appeared, the following vehicle (F-vehicle) was not yet in the drone videos. The following vehicle status variable(*F-val*) is set to 1. Relative position information(*dw*) is the headway distance between the S-vehicle and other vehicles, as shown in Figure 4. If the corresponding vehicle is not recorded in drone video, the corresponding *dw* is set to 0. Ultimately, a total of 54 indicators are taken as input variables. More details can be obtained from Table 1.

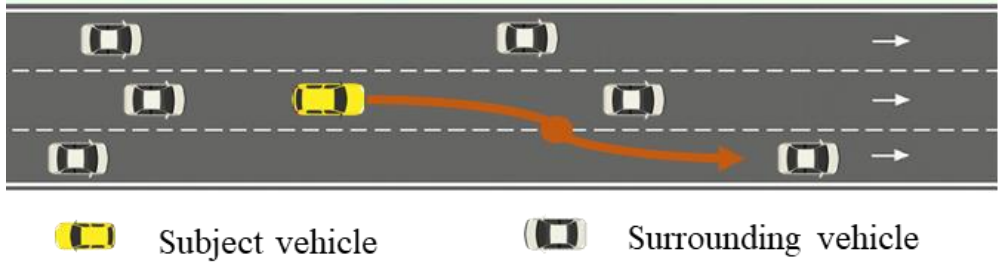

Fig.4 The headway distance between the S-vehicle and surrounding vehicles

TABLE 1

Input indicators of the model

| Inputs Variable | Variable descriptions |
| --- | --- |
| E-, P-, F-, LP-, LF-, RP-, RF-$v_x$ | The longitudinal velocity of E-vehicle and surrounding vehicle (ft/ sec) |
| E-, P-, F-, LP-, LF-, RP-, RF-$v_y$ | The lateral velocity of E-vehicle and surrounding vehicle (ft/ sec) |
| E-, P-, F-, LP-, LF-, RP-, RF-$a_x$ | The longitudinal acceleration of E-vehicle and surrounding vehicle (ft/ sec $^2$) |
| E-, P-, F-, LP-, LF-, RP-, RF-$a_y$ | The lateral acceleration of E-vehicle and surrounding vehicle (ft/ sec $^2$) |
| E-, P-, F-, LP-, LF -, RP-, RF-$\theta$ | The heading of E-vehicle and surrounding vehicle (degree) |

| E-, P-, F-, LP-, LF -, RP-, RF-$\Delta\theta$ | The yawRate of E-vehicle and surrounding vehicle (degrees/sec) |
|---|---|
| $dw_0, dw_1, dw_2, dw_3, dw_4, dw_5$ | Space headway between E-vehicle and surrounding vehicle (ft) |
| P-, F-, LP-, LF -, RP-, RF-val | 0 means it has recorded trajectory information; 1 means the trajectory information is missing |

Note: "E-" represents the ego vehicle; "P-" represents the closest preceding vehicle in the same lane; "F-" represents the closest following vehicle in the same lane; "LP-" represents the closest preceding vehicle in the adjacent left lane; "LF-" represents the closest following vehicle in the adjacent left lane; "RP-" represents the closest preceding vehicle in the adjacent right lane; "RF-" represents the closest following vehicle in the adjacent right lane;

**Method**

LC intention recognition is a multivariate time series classification problem. The indicators that require classification exhibit high-dimensionality. Selecting the appropriate model for this particular issue in machine learning applications can be a complex and challenging task. There are three main methods commonly used: Support vector machines (SVM), Ensemble Methods and Long Short-Term Memory model.

**4. 1 Support vector machines**

Support vector machine (SVM) is a supervised machine learning algorithm that is primarily used for classification tasks[5, 20]. An SVM performs classification by constructing a hyper-plane in higher dimensions. The main idea of SVM is to find an optimal hyperplane by mapping vectors to a higher-dimensional space. The hyperplane could effectively separate data points of different classes, and on either side of this separating hyperplane, two parallel hyperplanes are established.

**4.2 Ensemble Methods**

The ensemble method aims to improve the generalization and robustness performance of a single model by combining the results of multiple base estimators. eXtreme Gradient Boosting (XGBoost) and Light Gradient Boosting Machine (LightGBM) algorithms are two commonly used ensemble methods in machine learning.

(1) eXtreme Gradient Boosting algorithm

XGBoost is a machine learning algorithm that is based on Gradient Boosting Decision Trees first proposed by[26]. This algorithm encompasses an efficient linear model solver and regression tree algorithm, which can be applied in regression, and classification tasks. This technique combines multiple "weak" models to create a single "strong" model based on an additive training strategy. XGBoost combines regularization terms with a cost function to control the model complexity.

(2) LightGBM

LightGBM is a novel boosting framework proposed in 2017[27], which shares the same basic principle as XGBoost by using decision trees based on a learning algorithm. However, LightGBM has optimized the framework, mainly in terms of training speed.

**4.3 LSTM**

LSTM adopts a gating mechanism that selectively retains or forgets information, effectively enhancing the long-term dependency modeling capability of traditional RNNs. It could be employed individually to address such sequence-to-sequence prediction and time series classification issues. A typical LSTM block is configured mainly by an input gate $i_t$, forget gate $f_t$ and output gate $o_t$. These gates are computed as follows,

$$i_t = \sigma\left(W_i x_t + U_i h_{t-1} + b_i\right) \tag{8}$$

$$f_t = \sigma\left(W_f x_t + U_f h_{t-1} + b_f\right) \tag{9}$$

$$o_t = s\left(W_o x_t + U_o h_{t-1} + b_o\right) \tag{10}$$

Where $\sigma$ represents the sigmoid activation function; $x_t$ represents the input sequence at time $t$; $h_{t-1}$ represents the hidden state; $W$ is the parameter matrix at time $t$ and represents the input weight; U is the parameter matrix at time $t$-1 and represents the recurrent weight; $b_i, b_f$, and $b_o$ represent bias. The internal update state of the LSTM recurrent cells can be expressed as:

$$c_t = f_t \odot c_{t-1} + i_t \odot \tilde{c}_t \tag{11}$$

$$h_t = o_t \odot \tanh(c_t) \tag{12}$$

Where $\odot$ represents vector element-wise product, $c_t$ is the memory cell at time $t$-1, $\tilde{c}_t$ is the candidate memory at time t, $h_t$ is the outcome at time t.

*Evaluation indexes*

The modeling framework proposed includes classification models and sequence prediction models. The performance of classification models is evaluated from two aspects. One is the overall performance of the classification, and the other is the recognition performance of each class (Yang et al. 2021). The two indexes, precision and recall, are used to evaluate the detection performance of each class. The accuracy index measures the overall performance of the model. The three indexes can be calculated as follow,

$$Accuracy = \frac{T}{T + F} \qquad (13)$$

$$Precision = \frac{TP}{TP + FP} \qquad (14)$$

$$Rcall = \frac{TP}{TP + FN} \qquad (15)$$

Where T represents the number of correctly classified samples, F represents the number of incorrectly classified samples, TP is the number of correctly classified samples in a given class, FP is the number of incorrectly classified samples in a given class, FN denotes the number of incorrectly classified samples in a given class

**Results**

The parameter setting will affect the performance of the model. To obtain optimal parameter settings, some sensitivity experiments are performed on four models, using the control variable method. The parameters are selected based on the metrics of classification accuracy and training time. The final model used should minimize the training time of the model (reduce the complexity of the model) without compromising the accuracy of the model. With an equal number of samples, all experiments are conducted using the same device. The ten-fold cross-validation is adopted to train and test the model. As an example, the impact of the number of decision trees on classification accuracy was evaluated with maintaining the same input durations (input time duration = 5s). The dataset is randomly split into a training dataset and a test dataset with a ratio of 8:2. For training the LC intention classification model, eighty percent of total data is applied, and twenty percent of samples are used for testing the classification performance.

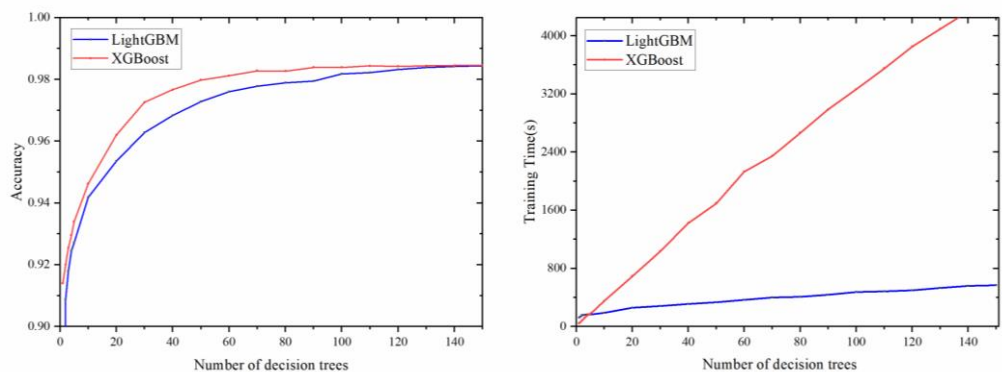

Fig.9 The impact of the decision trees number on recognition accuracy and training time

Fig.9 depicted the results of the experiment. It is evident that as the number of decision trees increases, the recognition accuracy and training time of the model will increase. The accuracy stabilizes once the number of decision trees reaches 120. The LightGBM algorithm exhibits a faster training speed comparing to XGBoost model. Similar results are obtained when utilizing sequences with different input durations. Hence, the number of decision tree for the LightGBM and XGBoost models are set to 120 in this study. To investigate the effect of input sequence length on classification outcomes, the performance of SVM, XGBoost, LightGBM, and LSTM models with varying input durations is evaluated. With an interval of 15 frames, a total of 12 input lengths are extracted from 30 frames(1s) to 180 frames(6s).

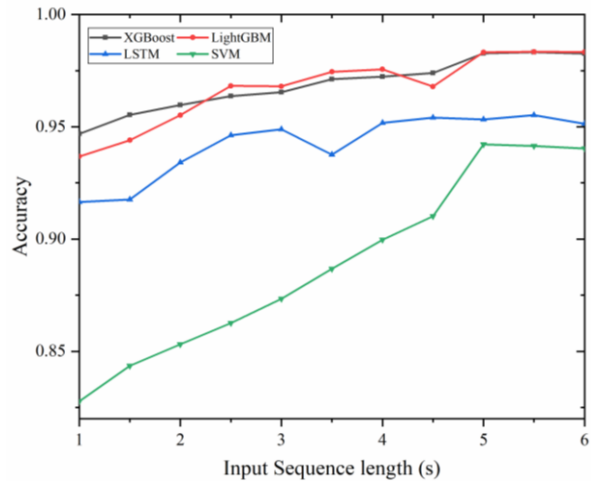

Fig.10 Accuracy comparison of LSTM, SVM, XGBoost, and LightGBM

Fig.10 illustrates the overall accuracy comparison results of the four models. It can observe that each model has good classification performance (above 80%), even though the four models are slightly different among different durations. With the same input data time scale, the two ensemble methods outperform SVM and LSTM models regarding classification accuracy. The best classification accuracy was achieved for three models (LSTM, XGBoost, and LightGBM) when the input length was five seconds. Despite not attaining optimal accuracy using a 5-second input time length, the SVM algorithm exhibited marginal enhancements in classification accuracy. Hence, a time duration $T = 150$ frames (5s) was chosen as input sequence lengths. Finally, 22160 RLC sequences and 15410 LLC sequences were extracted. To maintain data balance, 18000 LK sequences are randomly extracted from the raw dataset. Using the training dataset, the ten-fold cross-validated method is employed for model training and evaluation.

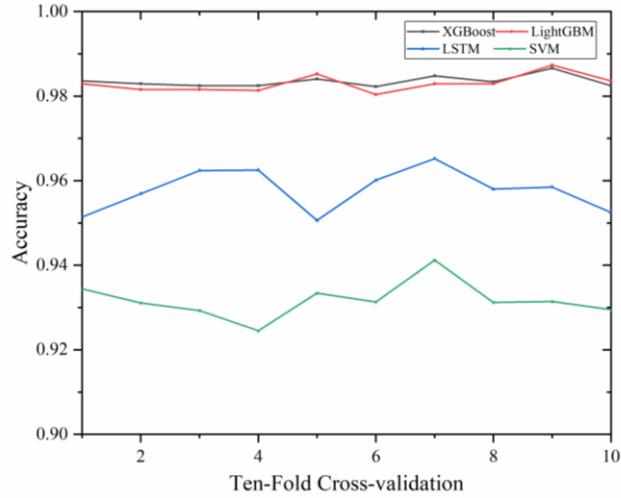

**Fig.11** Ten-fold Cross-validation for LSTM, SVM, XGBoost, and LightGBM

Fig.11 illustrates the outcomes of a Ten-fold cross-validation analysis conducted on LSTM, SVM, XGBoost, and LightGBM models. The average accuracy for LSTM and SVM is 0.9568 and 0.9317, with standard deviations of 0.007 and 0.004, respectively. XGBoost and LightGBM algorithms have average accuracies of 0.9835 and 0.9829, with a standard deviation of 0.001. The result indicated that XGBoost and LightGBM algorithms outperform LSTM and SVM regarding classification performance and demonstrate more stability. With an input length of 150 frames, Fig. 12 illustrates the confusion matrix for the four models using the validation set.

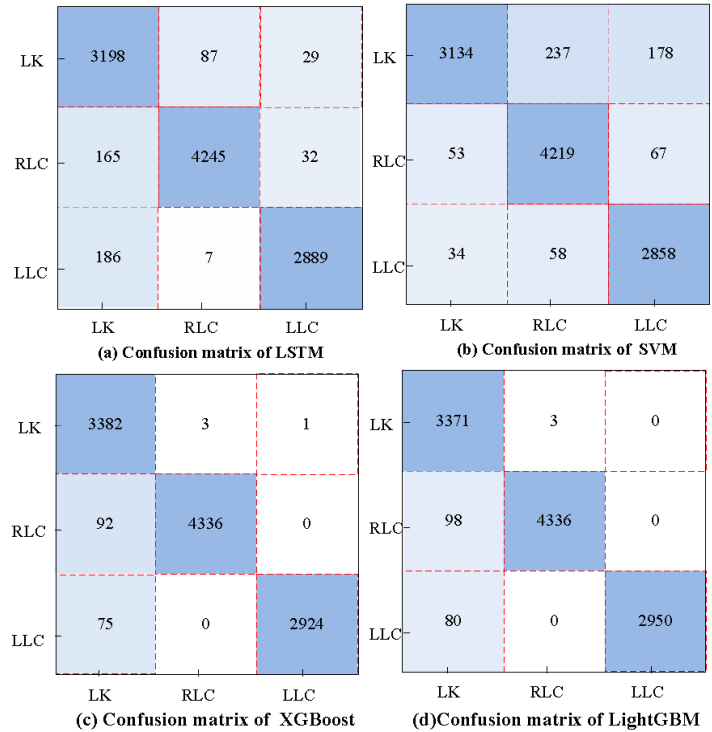

**Fig.12** Confusion matrix of classification models

Errors in classifying LC intentions can be categorized into three categories: the misidentification of LK as LC (Type I), the misclassification of LC as LK (Type II), and the misidentification of LLC and RLC from each other (Type III). Fig. 12 shows that XGBoost and LightGBM algorithms reduce the impact of Type II and Type III errors compared to LSTM and SVM models. Type I errors have a significant impact on the accuracy of all four models. This error could originate from two sources. One is that the model correctly identifies the behavior of a failed lane change. The other could be attributed to the variations in individual lane change behaviors among drivers[28, 29]. The LK process is influenced by factors such as driving style and driving ability, which can exceed the cognitive capabilities of the model, resulting in misjudgment. To provide a comprehensive assessment of classification performance, in addition to accuracy, other evaluation metrics such as precision, recall, and training time were evaluated through the confusion matrixes. The comparison results are displayed in Table 3.

Table 3 Evaluation results of LSTM, SVM, XGBoost, and LightGBM

| Model | Type | Precision | Recall | Accuracy | Training time(s) |
|---|---|---|---|---|---|
| LSTM | LK | 90.10% | 96.21% | 95.33% | 992.3 |
| | RLC | 97.83% | 95.78% | | |
| | LLC | 97.79% | 93.73% | | |
| SVM | LK | 88.31% | 97.29% | 94.21% | 33819.3 |
| | RLC | 97.23% | 93.46% | | |
| | LLC | 96.88% | 92.10% | | |
| XGBoost | LK | 95.29% | 99.88% | 98.42% | 3850.7 |
| | RLC | 99.93% | 97.92% | | |
| | LLC | 99.96% | 97.50% | | |
| LightGBM | LK | 99.91% | 94.98% | 98.32% | 496.4 |
| | RLC | 97.89% | 99.93% | | |
| | LLC | 97.34% | 100% | | |

The table demonstrates that the overall performance of SVM and LSTM is 94.21% and 95.33%, respectively. On the other hand, XGBoost and LightGBM algorithms achieve similar overall performances of 98.42% and 98.32%, respectively. The two **ensemble** algorithms outperform LSTM and SVM models and demonstrate a higher accuracy with an improvement of approximately 3.0%. With similar classification performance, the XGBoost algorithm requires six times more training time than the LightGBM algorithm. This result indicates that the LightGBM model provides a promising solution for driving intention classification tasks, as it outperforms other models in terms of both classification accuracy and computational efficiency.

**Discussion**

LC behavior is a fundamental driving operation that largely affects traffic efficiency and safety. In this paper, the LC vehicle status was characterized using six variables, including the longitudinal velocity (vx), lateral velocity (vy), longitudinal acceleration(ax), lateral acceleration (ay), vehicle heading ($\theta$), and yawRate ($\triangle\theta$). With using vehicle trajectory data, a novel ensemble method (LightGBM) was first utilized in this research.

For the LC intention recognition issues, this paper compared the classification performance of SVM, LSTM, XGBoost, and LightGBM models. The Ten-fold cross-validated method was employed for model training and evaluation. With an input length of 150 frames, the XGBoost and LightGBM models achieve an impressive overall classification performance of 98.42% and 98.32%, respectively. Compared to the LSTM and SVM models, the results demonstrate that two ensemble methods reduce the impact of Type I and Type III errors, demonstrating a higher accuracy with an improvement of approximately 3.0%. With approximately equal classification performance, it is noteworthy that the XGBoost algorithm necessitates six times more training time than the LightGBM algorithm.